\documentclass[11pt,a4paper]{article}
\usepackage[hyperref]{acl2019}
\usepackage{times}
\usepackage{latexsym}

\usepackage{url}

\aclfinalcopy 
\def\aclpaperid{532} 

\usepackage{graphicx}
\usepackage{subfigure}

\usepackage{amsmath,amsfonts,bm}

\newcommand{\topK}{\mathop{\mathrm{nearest\text{-}K}}}

\def\eqref#1{equation~\ref{#1}}

\def\1{\bm{1}}

\def\va{{\bm{a}}}

\def\vc{{\bm{c}}}

\def\vh{{\bm{h}}}

\def\vx{{\bm{x}}}
\def\vy{{\bm{y}}}

\DeclareMathAlphabet{\mathsfit}{\encodingdefault}{\sfdefault}{m}{sl}
\SetMathAlphabet{\mathsfit}{bold}{\encodingdefault}{\sfdefault}{bx}{n}

\DeclareMathOperator*{\argmax}{arg\,max}
\DeclareMathOperator*{\argmin}{arg\,min}

\usepackage{amsmath, amsthm, amssymb}
\usepackage{textcomp}
\usepackage{algorithm}
\usepackage{algorithmic}
\usepackage{multirow}
\usepackage{makecell}
\usepackage{graphicx}
\usepackage{stmaryrd}
\usepackage{upgreek}
\usepackage{bm}
\usepackage{cases}
\usepackage{mathtools}
\usepackage{rotating}
\usepackage{booktabs}
\usepackage{xcolor}
\usepackage{paralist}

\title{TIGS: An Inference Algorithm for Text Infilling with Gradient Search}

\author{
Dayiheng Liu\footnotemark[2], Jie Fu\footnotemark[3], Pengfei Liu\footnotemark[4], Jiancheng Lv\footnotemark[2]\hspace{1mm}\thanks{\hspace{2mm}Correspondence to Jiancheng Lv.} \\
\footnotemark[2]\hspace{0.5mm} College of Computer Science, Sichuan University\\
\footnotemark[3]\hspace{0.5mm} Mila, IVADO, Polytechnique Montreal\\
\footnotemark[4]\hspace{0.5mm} School of Computer Science, Fudan University\\
{\tt losinuris@gmail.com
 }
 \\
  {\tt lvjiancheng@scu.edu.cn}
  } 
\date{}
\begin{document}
\maketitle

\begin{abstract}
Text infilling is defined as a task for filling in the missing part of a sentence or paragraph, which is suitable for many real-world natural language generation scenarios. 
However, given a well-trained sequential generative model, generating missing symbols conditioned on the context is challenging for existing greedy approximate inference algorithms.
In this paper, we propose an iterative inference algorithm based on gradient search, which is the first inference algorithm that can be broadly applied to any neural sequence generative models for text infilling tasks.
We compare the proposed method with strong baselines on three text infilling tasks with various mask ratios and different mask strategies. 
The results show that our proposed method is effective and efficient for fill-in-the-blank tasks, consistently outperforming all baselines.\footnote{Our code and data are available at \url{https://github.com/dayihengliu/Text-Infilling-Gradient-Search}}
\end{abstract}

\section{Introduction}
\label{sec:intro}

Text infilling aims at filling in the missing part of a sentence or paragraph by making use of the past and future information around the missing part, which can be used in many real-world natural language generation scenarios, for example, fill-in-the-blank image captioning \cite{Sun2017}, lexically constrained sentence generation \cite{liu2018bfgan}, missing value reconstruction (e.g. for damaged or historical documents) \cite{Berglund2015}, acrostic poetry generation \cite{liu2018multi}, and text representation learning \cite{devlin2018bert}.

\begin{figure}[t] 
   \centering
   \includegraphics[scale=0.43]{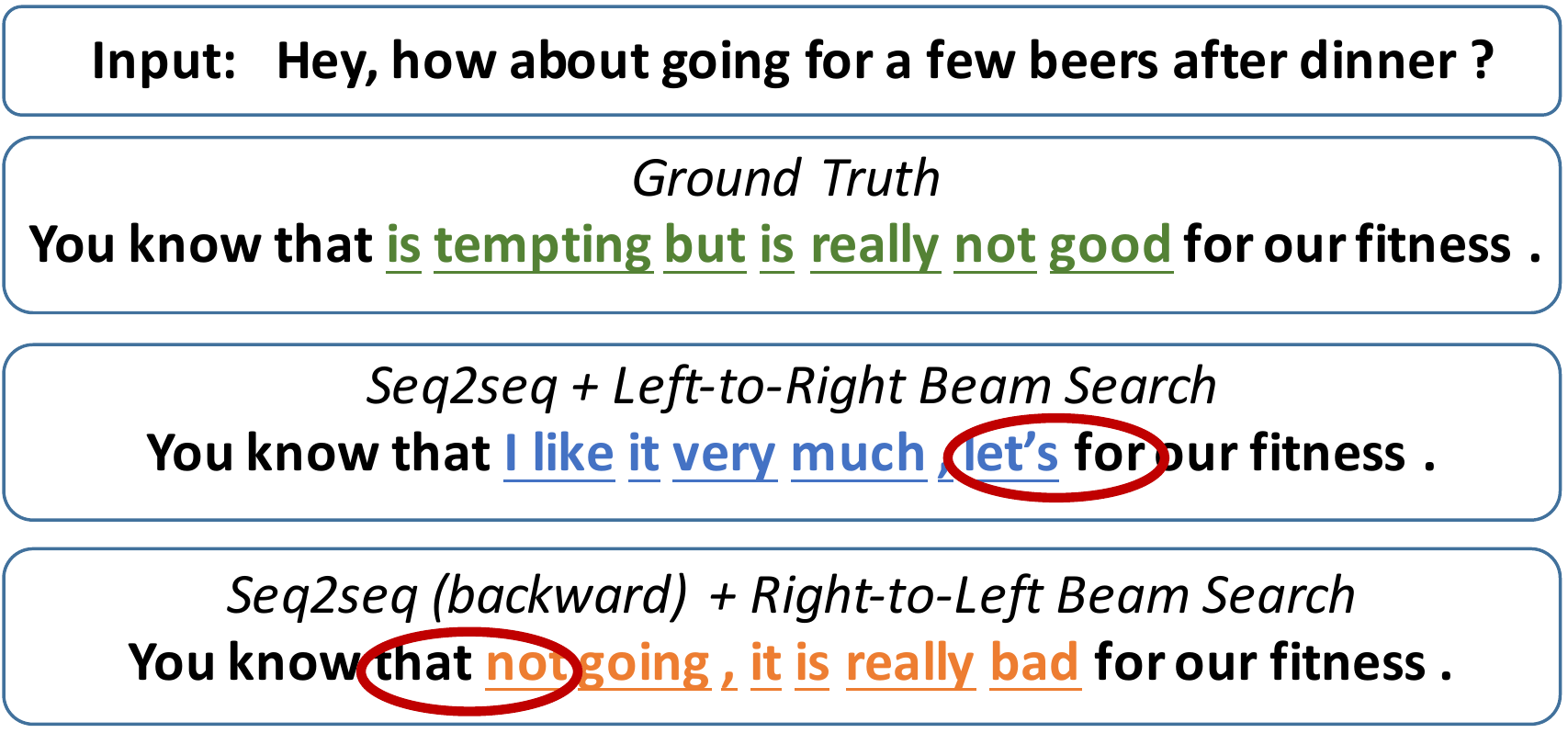}
   \vspace{-5pt}
   \caption{Our key observation on text infilling for Dialogue task. The inability of unidirectional BS to consider both the future and past contexts leads models to fill the blank with words that clash abruptly with the context around the blanks (see red circles).}
   \label{fig:intro}
   \vspace{-10pt}
\end{figure}

Text infilling is an under-explored challenging task in the field of text generation. Recently, sequence generative models like sequence-to-sequence (seq2seq) models \cite{sutskever2014sequence,bahdanau2014neural,gehring2017convolutional,vaswani2017attention} are widely used in text generation tasks, including neural machine translation \cite{wu2016google,vaswani2017attention}, image captioning \cite{Anderson2017Guided}, abstractive summarization \cite{See2017Get}, and dialogue generation \cite{Mei2016Coherent}. Unfortunately, given a well-trained\footnote{Here ``well-trained'' means that ones focus on popular model settings and data sets, and follow standard training protocols.} 
neural seq2seq model or unconditional neural language model \cite{mikolov2010recurrent}, it is a daunting task to directly apply it to text infilling task. As shown in Figure~\ref{fig:intro}, we observe that the infilled words should be conditioned on past and future information around the missing part,  which is contrary to the popular learning paradigm, namely, each output symbol is conditioned on all previous outputs during inference by using unidirectional Beam Search (BS) \cite{och2004alignment}. 

To solve the issues above, one family of methods for text infilling is ``trained to fill in blanks'' \cite{Berglund2015,fedus2018maskgan,zhu2019text}, which requires large amounts of data in fill-in-the-blank format to train a new model that takes the output template as a conditional input. 
Such methods are only used for unconditional text infilling tasks, whereas many text infilling tasks are conditional, e.g., conversation reply with templates. 
Another kind of promising approach \cite{Berglund2015,wang2016image,Sun2017} is an inference algorithm that can be directly applied to other generative models. 
These inference algorithms are applied to Bidirectional RNNs (BiRNNs) \cite{schuster1997bidirectional,baldi1999exploiting} which can model both forward and backward dependencies.
The latest work is Bidirectional Beam Search (BiBS) \cite{Sun2017} which proposes an approximate inference algorithm in BiRNN for image caption infilling.
However, this method is based on some unrealistic assumptions, such as that given a token, its future sequence of words is independent of its past sequence of words.
We experimentally find that these assumptions often generate non-smooth or unreal complete sentences.
Moreover, these inference algorithms can be only used to decoders with bidirectional structures, whereas almost all sequence generative models use a unidirectional decoder. As a result, it is highly expected to develop an inference algorithm that could be applied to the unidirectional decoder.

In this paper, we study the general inference algorithm for text infilling to answer the question: 
\begin{itemize}
\setlength{\itemsep}{0pt}
\setlength{\parsep}{0pt}
\setlength{\parskip}{0pt}
  \item \textit{Given a well-trained neural sequence generative model, is there any inference algorithm that can effectively fill in the blanks in the output sequence?}
\end{itemize}

To investigate such a possibility, we propose a dramatically different inference approach called Text Infilling with Gradient Search (TIGS), in which we search for infilled words based on gradient information to fill in the blanks. To the best of our knowledge, \textbf{this could be the first inference algorithm} that does not require any modification or training of the model and can be broadly used in any sequence generative model to solve the fill-in-the-blank tasks as verified in our experiments.

To be specific, we treat the blanks to be filled as parameterized vectors during inference. More concretely, we first randomly or heuristically project each blank to a valid token and initialize its parameterized vector with the word embedding of the valid token. The goal is seeking the words to be infilled by minimizing the negative log-likelihood (NLL) of the complete sequence.
Then the algorithm alternately performs optimization step (\textbf{O-step}) and projection step (\textbf{P-step}) until convergence.
In \textbf{O-step}, we fix all other parameters of the model and only optimize the blank parameterized vectors by gradients. 
In \textbf{P-step}, heuristics like local search and projected gradient are used to project the blank parameterized vectors to valid tokens (i.e., discretization). 

The contribution and novelty of this work could be summarized as below:
\begin{itemize}
\setlength{\itemsep}{0pt}
\setlength{\parsep}{0pt}
\setlength{\parskip}{0pt}
  \item We propose an iterative inference algorithm based on gradient search, which could be the first inference algorithm that can be broadly applied to any neural sequence generative models for text infilling tasks.
  \item Extensive experimental comparisons show the effectiveness and efficiency of the proposed method on three different text infilling tasks, compared with five state-of-the-art methods.
\end{itemize}

\section{Related Works}
\label{sec:related}

There are some effective solutions to the text infilling task: a) training a model specifically for text infilling tasks \cite{Berglund2015,fedus2018maskgan,zhu2019text}; b) using standard sequence generative model with modified inference algorithm \cite{Berglund2015,wang2016image,Sun2017}. 

As one typical work of the first category, NADE \cite{Berglund2015} is proposed to train a specific BiRNNs for filling in blanks, which concatenates an auxiliary vector to input vectors for indicating a missing input during training and inference. 
\newcite{fedus2018maskgan} propose MaskGAN which 1) uses some specific ``missing'' tokens to indicate the blanks and takes the whole sequence with blanks (called template) as the input of encoder, and 2) uses an RNN as a decoder to generate the whole sentence after filling in the blanks.
Similarly, \newcite{zhu2019text} use self-attention model \cite{vaswani2017attention}, which takes the template as the input for unconditional text infilling task.
One major limitation of these works is that they require large amounts of data in fill-in-the-blank format and need to train a specific model. 
Besides, they are only used for unconditional text infilling tasks. Different from them, our new inference algorithm does not require any modification or training of the model, which can be broadly applied to any neural seq2seq models for both conditional and unconditional text infilling tasks.

As with the second category, some inference algorithms based on BiRNNs have been proposed for fill-in-the-blank tasks thanks to their ability to model both forward and backward dependencies. For example, 
\newcite{Berglund2015} propose Generative Stochastic Networks (GSN) to reconstruct the blanks of sequential data. 
The idea is to first randomly initialize the symbols in the blanks and then resample an output $y_t$ from $\mathrm{P}_{BiRNN}(y_t \mid \{y_d\}_{d \neq t}, x)$ one at a time until convergence.
More recently, \newcite{Sun2017} propose the Bidirectional Beam Search (BiBS) inference algorithm of BiRNNs for fill-in-the-blank image captioning task.
However, this method is based on some strong assumptions, which may be violated in practice.
As shown in our experiments, we provide empirical analysis on cases where this approach fails. 
Moreover, GSN and BiBS can be only applied to decoders with bidirectional structures, while almost all sequence generative models use a unidirectional decoder.
In contrast, our proposed inference method decouples from these assumptions and can be applied to the unidirectional decoder. 

\begin{figure*}[t] 
   \centering
   \includegraphics[width=4.0in]{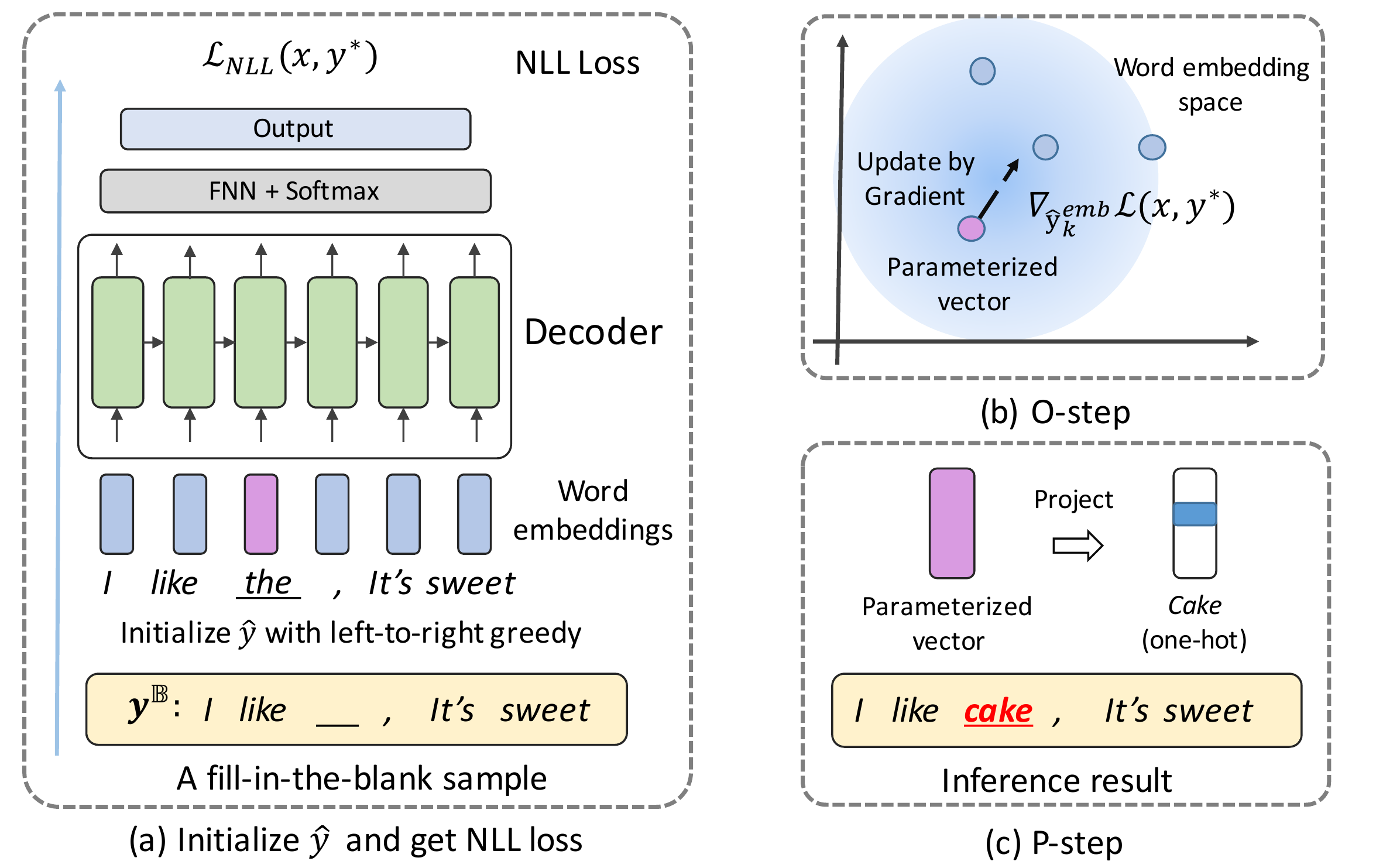}
   \caption{Overall framework.}
   \label{fig:overview}
\end{figure*}

\section{Preliminary}
\label{sec:preliminary}
Since our method utilizes gradient information, it could smoothly cooperate with other architectures, such as models proposed in \citep{vaswani2017attention,gehring2017convolutional}. Considering the popularity of RNNs, and the related work is based on RNN model, we use RNN-based models as a showcase in this paper.

\subsection{RNN-based Seq2Seq Model}
We firstly introduce the notations and briefly describe the standard RNN-based seq2seq model.
Let $\vx=\{\vx_1, \vx_2 , ... , \vx_n\}$ denotes one-hot vector representations of the conditional input sequence, $\vy=\{y_1, y_2, ... , y_m\}$ denotes scalar indices of the corresponding target sequence, and $\mathcal{V}$ denotes the vocabulary. $n$ and $m$ represent the length of the input sequence and the output sequence, respectively.

The seq2seq model is composed of an encoder and a decoder. For the encoder part, each $\vx_t$ will be firstly mapped into its corresponding word embedding $\vx^{emb}_t$. 
Then $\{\vx^{emb}_t\}$ are input to a bi-directional or unidirectional long short-term memory (LSTM) \cite{hochreiter1997long} RNN to get a sequence of hidden states $\{\vh^{enc}_t\}$. 

For the decoder, at time $t$, similarly $y_t$ is first mapped to $\vy^{emb}_t$. Then a context vector $\vc_t$ is calculated with attention mechanism \cite{bahdanau2014neural, luong2015effective} $\vc_t=\sum^n_{i=1}\va_{ti}\vh^{enc}_i$, which contains useful latent information of the input sequence. Here, $\va_{t}$ is an attention distribution vector to decide which part to focus on. The context vector $\vc_t$ and the embedding $\vy^{emb}_t$ are fed as input to a unidirectional RNN language model (LM), which will output a probability distribution of the next word $P(y_{t+1} | \vy_{1:t}, \vx)$, where $\vy_{1:t}$ refers to $\{y_1,...,y_t\}$.

During training, the negative log-likelihood (NLL) of the target sequence is minimized using standard maximum-likelihood (MLE) training with stochastic gradient descent (SGD), where the NLL is calculated as follows:
\begin{equation}
\setlength{\abovedisplayskip}{5pt}
\setlength{\belowdisplayskip}{5pt}
-\log P(\vy|\vx) = - \sum^m_{t=1} \log P(y_t|\vy_{1:t-1}, \vx).
\end{equation}

During inference, the decoder needs to find the most likely output sequence $\vy^*$ by giving the input $\vx$:
\begin{equation}
\setlength{\abovedisplayskip}{5pt}
\setlength{\belowdisplayskip}{5pt}
\vy^* = \mathop{\argmax}_{\vy} P(\vy|\vx). 
\end{equation}

Since the number of possible sequences grows as $|\mathcal{V}|^{m}$ ($|\mathcal{V}|$ is the size of vocabulary), exact inference is NP-hard and approximate inference algorithms like left-to-right greedy decoding or beam search (BS) \cite{och2004alignment} are commonly used.
 
\subsection{Problem Definition}
In this paper, instead of setting some restrictions such as limiting the number of blanks or restricting the position of the blanks in previous work \cite{Sun2017}, we consider a more general case of text infilling task where the number and location of blanks are arbitrary. 

Let $\underline{B}$ be a placeholder for a blank, $\mathbb{B}$ be the set that records all the blanks' position index, and $\vy^{\mathbb{B}}$ be a target sequence where portions of text body are missing as indicated by $\mathbb{B}$. For instance, if a target sequence has two blanks at the position $i$ and $j$, then $\mathbb{B}=\{i, j\}$ and $\vy^{\mathbb{B}}=\{y_1,..,y_{i-1},\underline{B}, y_{i+1}, ..., y_{j-1},\underline{B}, y_{j+1},..., y_m \}$.

Given an input sequence $\vx$ and a target sequence $\vy^{\mathbb{B}}$ containing blanks indicated by $\mathbb{B}$, we aim at filling in the blanks of $\vy^{\mathbb{B}}$.
This procedure needs to consider the global structure of sentences and provide meaningful details under the condition $\vx$.

\section{Methodology}
\label{sec:methodology}
In this section, we present our inference method in detail. 
The overall framework is shown in Figure \ref{fig:overview}. 
Given a well-trained seq2seq model and a pair of text infilling data ($\vx$, $\vy^{\mathbb{B}}$), the method aims at finding an infilled word set $\hat{\vy} =\{\hat{y}_1,...,\hat{y}_{|\mathbb{B}|}\}$ to minimize the NLL of the complete sentence $\vy^*$ via:
\begin{equation}
\setlength{\abovedisplayskip}{5pt}
\setlength{\belowdisplayskip}{5pt}
    \hat{\vy} = \argmin \limits_{\hat{y}_j \in \mathcal{V}} \mathcal{L}_{NLL}(\vx,\vy^*),
\end{equation}
where $\vy^*$ denotes the complete sentence after filling the blanks of $\vy^{\mathbb{B}}$ with $\hat{\vy}$, $|\mathbb{B}|$ denotes the number of blanks.
Since the number of possible infilled word set is $|\mathcal{V}|^{|\mathbb{B}|}$, na\"ively searching in this space is NP-hard. 

Our key idea is to utilize the gradient information to narrow the range of search during inference. This idea is similar to the ``white-box'' adversarial attacks \cite{goodfellow6572explaining,szegedy2013intriguing,he2018detecting}.
However, the adversarial attack aims to slightly modify the inputs in order to mislead the model to make wrong predictions, while our goal is to search for the reasonable words that should be filled into the blanks.

Unlike the continuous input space (e.g., images) in other tasks, applying gradient search directly to the input would make it invalid (i.e., no longer a one-hot vector) for text infilling tasks. 
More specifically, we firstly treat the blanks to be filled as parameterized word embedding vectors $\hat{\vy}^{emb} = \{\hat{\vy}_1^{emb},...,\hat{\vy}_{|\mathbb{B}|}^{emb}\}$. Then, we fix the parameters of the well-trained model and only optimize these parameterized vectors in the continuous space, where the gradient information can be used to minimize the NLL loss $\mathcal{L}_{NLL}(\vx,\vy^*)$. 
Finally, the $\hat{\vy}^{emb}$ is discretized into valid words $\hat{\vy}$ by measuring the distance between the $\hat{\vy}^{emb}$ and the word embeddings in $\mathbb{W}^{emb}$. Here $\mathbb{W}^{emb}$ denotes the word embedding matrix in the decoder of the well-trained seq2seq model, and each column of $\mathbb{W}^{emb}$ represents the word embedding of one word in the vocabulary.

As every word in the set $\hat{\vy}$ is dependent on each other, the simultaneous discretization of all parameterized word embeddings in $\hat{\vy}^{emb}$ into valid words at the same time usually make the complete sentence $\vy^*$ non-smooth.
As a concrete example, when infilling the two blanks in ``Amy likes eating \underline{\quad} \underline{\quad}, so she goes to snack bars very often.'', the $\hat{\vy}^{emb}$ may be close to the word embeddings of \{``\underline{ice}'', ``\underline{cream}''\} and \{``\underline{fried}'', ``\underline{chips}''\}. 
However, if one discretizes the two blanks simultaneously, one might get answers like \{``\underline{ice}'', ``\underline{chips}''\} or \{``\underline{fried}'', ``\underline{cream}''\}. 
Therefore, we adopt an iterative algorithm which is similar to Gibbs sampling. 
At each inference step, we focus on one single infilled word $\hat{y}_j$ for $j$-th blank and update it while keeping other words in the infilled word set $\hat{\vy}$ fixed. For the unknown blank length tasks (each blank may contain an arbitrary unknown number of tokens), we can apply the TIGS as a black box inference algorithm over a range of blank lengths and then rank these solutions.

At the beginning, we initialize the infilled word set $\hat{\vy}$ with some valid words randomly or heuristically (from a left-to-right beam search). 
Then we perform optimization step (\textbf{O-step}) and projection step (\textbf{P-step}) alternately to update each infilled word in the infilled word set $\hat{\vy}$ until convergence or reach the maximum number of rounds $T$.

In \textbf{O-step}, we aim to optimize the $\hat{\vy}^{emb}_j$ in continuous space using gradient information with respect to $\mathcal{L}_{NLL}(\vx,\vy^*)$.
Firstly, we get the complete sentence $\vy^*$ by filling $\hat{\vy}$ in the blanks of $\vy^{\mathbb{B}}$ and obtain the $\mathcal{L}_{NLL}(\vx,\vy^*)$ of $\vy^*$ after putting $\vx$ into the encoder and $\vy^*$ into the decoder of the well-trained seq2seq model.
Then we treat the vector $\hat{\vy}^{emb}_j$ as parameterized vector, and fix all other parameters of the seq2seq model and only optimize the parameterized vector $\hat{\vy}_j^{emb}$ with gradient information to minimize $\mathcal{L}_{NLL}(\vx,\vy^*)$.

However, directly optimizing $\mathcal{L}_{NLL}(\vx,\vy^*)$ may lead to the final $\hat{\vy}^{emb}_j$ not like a feasible word embedding in $\mathbb{W}^{emb}$, and its nearest neighbor word embedding in $\mathbb{W}^{emb}$ could be far away from it.
So we add an $L$2 penalty to make the $\hat{\vy}^{emb}_j$ get close to $\mathbb{W}^{emb}$:
\begin{equation}
\setlength{\abovedisplayskip}{5pt}
\setlength{\belowdisplayskip}{5pt}
    \mathcal{L}(\vx,\vy^*) = \mathcal{L}_{NLL}(\vx,\vy^*) + \lambda \cdot \sum_j \left\|\hat{\vy}^{emb}_j\right\|_2,
\end{equation}
where $\lambda$ is a hyperparameter. We also tried to add an additional regularization term that directly narrow the distance between $\hat{\vy}^{emb}_j$ and its nearest word embedding in $\mathbb{W}^{emb}$, which is used in \newcite{cheng2018seq2sick} for seq2seq adversarial attacks, but no obvious improvement was found.

Given the loss $\mathcal{L}(\vx,\vy^*)$, $\hat{\vy}^{emb}_j$ is updated with $\nabla_{\hat{\vy}_j^{emb}}\mathcal{L}(\vx,\vy^*)$ by one-step gradient descent:
\begin{equation}
\setlength{\abovedisplayskip}{5pt}
\setlength{\belowdisplayskip}{5pt}
\hat{\vy}_j^{emb} = \hat{\vy}_j^{emb} - \alpha \cdot \nabla_{\hat{\vy}_j^{emb}}\mathcal{L}_{NLL}(\vx,\vy^*).
\end{equation}

Instead of updating $\hat{\vy}^{emb}_j$ by na\"ively SGD algorithm, we experimentally find that  Nesterov \cite{SutskeverMDH13} optimizer performed better than  other optimizers to update $\hat{\vy}_j^{emb}$. 
As discussed in \newcite{dong2017boosting}, this momentum based optimizer can stabilize update directions and escape from poor local maxima during the iterations for adversarial attack.

In \textbf{P-step}, we aim to project the $\hat{\vy}_j^{emb}$ into a valid infilled word $\hat{y_j}$.
A na\"ive way is to find the word whose word embedding in $\mathbb{W}^{emb}$ is nearest to $\hat{\vy}_j^{emb}$ based on the distance metric function $dist(\cdot)$\footnote{Through experiments we find that using Euclidean distance as metric function $dist(\cdot)$ perform slightly better than Cosine distance for our method.}.
However, due to its high dimensionality, the obtaining word embedding may be far from satisfactory.  
Instead, similar to the idea of beam search, we first obtain a set $S$ containing $K$ candidate words whose word embedding is K-nearest to $\hat{\vy}_j^{emb}$:
\begin{equation}
\setlength{\abovedisplayskip}{5pt}
\setlength{\belowdisplayskip}{5pt}
S = \topK \limits_{y_k \in \mathcal{V}} dist(\hat{\vy}_j^{emb},\vy^{emb}_k),
\end{equation}
and then we select one word with lowest NLL from these $K$ words in $S$ as $\hat{y}_j$.
Our experiments suggest that just setting the size of $K$ to 1\% of the vocabulary size works well.

The whole algorithm is further summarized in \ref{alg:1}. Since our method is designed for the unidirectional decoder, the time complexity is expected to be slightly higher than that of the inference algorithm designed for the bidirectional decoder. In brief, our approach requires $mKT|\mathbb{B}|$ RNN steps, while the GSN \cite{Berglund2015} requires $mT|\mathbb{B}|$ BiRNN steps, and the BiBS \cite{Sun2017} requires $2mKT$ RNN steps. Fortunately, our inference algorithm can be easily optimized with GPUs.

\begin{algorithm}[th] \small
   \caption{TIGS algorithm}
   \label{alg:1}
    \begin{algorithmic}
       \STATE {\bfseries Input:} a trained seq2seq model, a pair of text infilling data ($\vx$, $\vy^{\mathbb{B}}$), output length $m$.
       \STATE {\bfseries Output:} a complete output sentence $\vy^*$.
       \STATE Initialize the infilled word set $\hat{\vy}$ and initialize $\vy^*$ by infilling $\vy^{\mathbb{B}}$ with $\hat{\vy}$.
       \STATE Initialize $\hat{\vy}^{emb}$ by looking up the word embedding matrix $\mathbb{W}^{emb}$.
       \FOR{$t=1,2,\dots, T$}
          \FOR{$j=1,2,\dots,|\mathbb{B}|$}
            \STATE \textbf{O-step}:
             \STATE Update $\hat{\vy}_j^{emb}$ with gradient $\nabla_{\hat{\vy}_j^{emb}}\mathcal{L}(\vx,\vy^*)$ 
            \STATE \textbf{P-step}: 
            \STATE Set $S = \topK \limits_{y_k \in \mathcal{V}} dist(\hat{\vy}_j^{emb},\vy^{emb}_k)$
            \STATE Set $\hat{y}_j = \argmin \limits_{\hat{y}_j \in S} \mathcal{L}_{NLL}(\vx,\vy^*)$
           \ENDFOR
           \STATE Update $\vy^*$ with $\hat{y}_j$
       \IF{convergence} 
       \STATE {\bfseries break} 
       \ENDIF
       \ENDFOR
       \STATE {\bfseries return} $\vy^*$
    \end{algorithmic}
\end{algorithm}
\section{Experiments}
\label{sec:experiments}

\subsection{Datasets}
In the experiments, we evaluate the proposed method on three text infilling tasks with three widely used publicly available corpora.

The first task is \textbf{conversation reply with a template} (denoted as Dialog) which is conducted on the DailyDialog \cite{LiSSLCN17} dataset. 
We use its single-turn data, which contains 82,792 conversation pairs.
The query sentence is taken as encoder input $\vx$, and the reply sentence is taken as $\vy$.

The second task is \textbf{Chinese acrostic poetry generation} (denoted as Poetry).
Here we use a publicly available Chinese poetry dataset\footnote{\url{https://github.com/chinese-poetry/chinese-poetry}} which contains 232,670 Chinese four-line poems. 
For each poem, the first two lines are used as encoder input $\vx$, and the last two lines are $\vy$.

The third task is \textbf{infilling product reviews} (denoted as APRC). 
The Amazon Product Reviews Corpus (APRC) \cite{Dong2017Learning}, which is built upon Amazon product data \cite{mcauley2015inferring} and contains 347,061 reviews, is used in this task. 
Unlike the first two tasks, this task is an unconditional text infilling task (without conditional input $\vx$).
We use each product review in \newcite{Dong2017Learning} as $\vy$.

\begin{figure}[t] 
   \centering
   \includegraphics[width=3.0in]{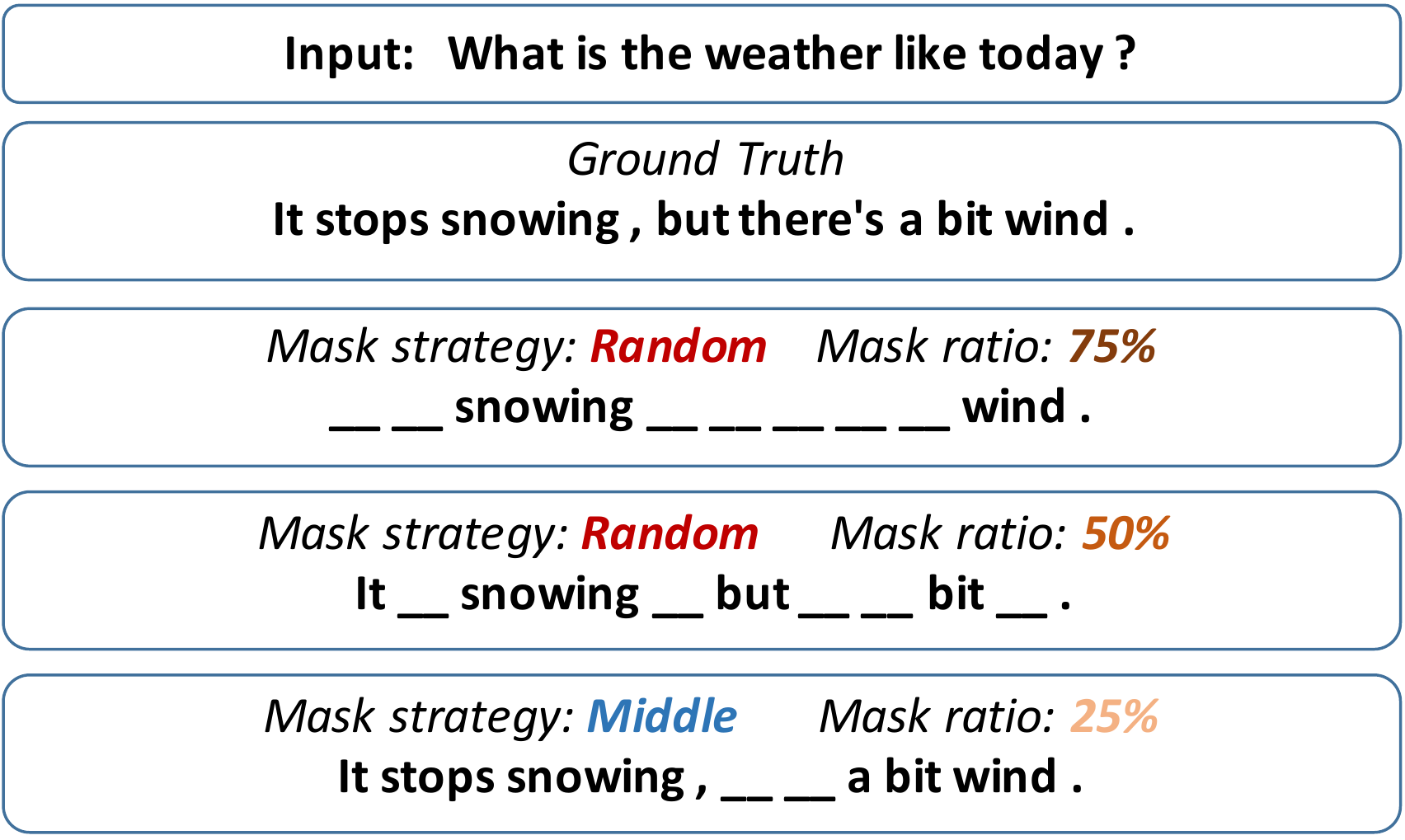}
      \vspace{-15pt}
   \caption{Some testing samples of conversation reply with templates task with different mask strategies and ratios.}
      \vspace{-10pt}
   \label{fig:data_example}
\end{figure}

For each task, we take 5,000 samples in the test set to construct the data with blanks ($\vy^{\mathbb{B}}$) for testing, we create a variety of test samples by masking out text $\vy$ with varying missing ratios and two mask strategies. More specifically, the first mask strategy is called \textbf{middle} which is followed as the setting in \newcite{Sun2017}, namely, removing $r = 25\%$, $50\%$, or $75\%$ of the words from the middle of $\vy$ for each data. 
The second mask strategy is called \textbf{random}, namely, randomly removing $r = 25\%$, $50\%$, or $75\%$ of the words in $\vy$ for each data.
To sum up, we have three test tasks, and each task has six types of test sets (two mask strategies and three mask ratios). 
Each test set contains 5,000 test samples.
We show some data examples in Figure \ref{fig:data_example}.

\subsection{Baselines}
We compare our approach \textbf{TIGS} with several strong baseline approaches: 

\textbf{Seq2Seq-f}: it runs beam search (BS) with beam width $K$ on a well-trained seq2seq model (forward) to fill the blanks from left to right.

\textbf{Seq2Seq-b}: it runs BS with beam width $K$ on a well-trained seq2seq model (backward) to fill the blanks from right to left.

\textbf{Seq2Seq-f+b}: it fills the blanks by both Seq2Seq-f and Seq2Seq-b, and then selects the output with a maximum of the probabilities assigned by the seq2seq models. 
This method is used in \newcite{wang2016image}. 

\textbf{BiRNN-GSN}: it runs GSN \cite{Berglund2015} on a well-trained seq2seq model with BiRNN as the decoder to fill the blanks. 

\textbf{BiRNN-BiBS}: it runs bidirectional beam search (BiBS) \cite{Sun2017} on a well-trained seq2seq model with BiRNN as the decoder to fill the blanks. The method has achieve the state-of-the-art results on fill-in-the-blank image captioning task in \newcite{Sun2017}.

Except for BiRNN-GSN and BiRNN-BiBS, all the above baselines and our method perform inference on the same well-trained seq2seq model.
BiRNN-GSN and BiRNN-BiBS perform inference on a well-trained seq2seq model in which the decoder is BiRNN.
These models are trained on the complete sentence dataset with standard maximum-likelihood. 
Moreover, the sentences with blanks are only used in the inference stage.
For fari comparison, BiRNN-BiBS, BiRNN-GSN, and the proposed method use the same initialization strategy (left-to-right greedy). The maximum number of iterations $T$ is set to 50 to ensure that all the algorithms can achieve their best performance.

In addition to the above inference based approaches, we also compare two model-based approaches: \textbf{Mask-Seq2Seq} and \textbf{Mask-Self-attn} \cite{fedus2018maskgan,zhu2019text}. 
These baselines take the output template as an additional input and are trained on the data in fill-in-the-blank format.
We use LSTM RNNs for Mask-Seq2Seq, and use the self-attention model \cite{vaswani2017attention} for Mask-Self-attn \cite{zhu2019text} which is shown to have better performance than GAN-based models \cite{Goodfellow14} for text infilling.

\begin{table*}[ht]  \scriptsize 
\centering
\setlength{\tabcolsep}{6pt}
\begin{center}
\begin{tabular}{c c l c c c c c c c c c c c c c c c c c}
\toprule
\multirow{2}{*}{Datasets} & \multirow{2}{*}{Metrics} & \multirow{2}{*}{Methods} & \multicolumn{2}{c}{$r$=25\%} & \multicolumn{2}{c}{$r$=50\%} & \multicolumn{2}{c}{$r$=75\%} \\
 \cmidrule(lr){4-5}
  \cmidrule(lr){6-7}
  \cmidrule(lr){8-9}
& & & Random & Middle & Random & Middle & Random & Middle \\
\midrule
\multirow{17}{*}{Dialog}
& \multirow{8}{*}{NLL}
& Seq2Seq-f      & 3.573  & 3.453 & 3.653 & 3.316 & 3.328 & 2.975  \\
& & Seq2Seq-b      & 3.657  & 3.558 & 3.911 & 3.542 & 3.713 & 3.421  \\
& & Seq2Seq-f+b    & 3.397  & 3.321 & 3.491 & 3.213 & 3.233 & 2.932  \\
& & BiRNN-BiBS     & 3.248  & 3.279 & 3.268 & 3.294 & 3.245 & 3.217  \\ 
& & BiRNN-GSN      & 3.239  & 3.270 & 3.219 & 3.199 & 3.086 & 2.938  \\ 
& & Mask-Seq2Seq   & 3.406  & 3.368 & 3.434 & 3.347 & 3.279 & 3.177  \\
& & Mask-Self-attn & 3.567  & 3.524 & 3.694 & 3.466 & 3.509 & 3.205  \\
& & TIGS (ours)  & \textbf{3.143} & \textbf{3.164} & \textbf{3.050} & \textbf{3.030} & \textbf{2.920} & \textbf{2.764}  \\
\cmidrule(lr){2-9}
& \multirow{9}{*}{BLEU}
& Template       & 0.780  & 0.823 & 0.621 & 0.700 & 0.552 & 0.601  \\
& & Seq2Seq-f      & 0.834  & 0.861 & 0.670 & 0.737 & 0.584 & 0.640  \\
& & Seq2Seq-b      & 0.837  & 0.862 & 0.675 & 0.739 & 0.584 & 0.627  \\
& & Seq2Seq-f+b    & 0.860  & 0.881 & 0.692 & 0.751 & 0.594 & 0.643  \\
& & BiRNN-BiBS     & 0.828  & 0.852 & 0.661 & 0.725 & 0.575 & 0.626  \\ 
& & BiRNN-GSN      & 0.894  & 0.892 & \textbf{0.726} & 0.752 & 0.600 & 0.643 \\ 
& & Mask-Seq2Seq   & 0.867  & 0.887 & 0.719 & 0.769 & 0.614 & \textbf{0.662} \\
& & Mask-Self-attn & 0.858  & 0.864 & 0.719 & 0.743 & \textbf{0.623} & 0.643 \\
& & TIGS (ours)      & \textbf{0.895} & \textbf{0.894} & 0.724 & \textbf{0.754} & 0.596 & 0.644 \\

\midrule
\multirow{17}{*}{Poetry}
& \multirow{8}{*}{NLL}
   & Seq2Seq-f      & 4.107 & 4.022 & 3.901 & 3.642 & 3.430 & 3.294  \\
&  & Seq2Seq-b      & 4.180 & 4.124 & 4.051 & 3.837 & 3.638 & 3.511  \\
&  & Seq2Seq-f+b    & 4.021 & 3.994 & 3.825 & 3.630 & 3.390 & 3.275  \\
&  & BiRNN-BiBS     & 3.939 & 3.966 & 3.735 & 3.701 & 3.476 & 3.430  \\ 
&  & BiRNN-GSN      & 3.953 & 3.976 & 3.739 & 3.652 & 3.405 & 3.296  \\
&  & Mask-Seq2Seq   & 4.103 & 4.071 & 3.996 & 3.886 & 3.738 & 3.637  \\
&  & Mask-Self-attn & 4.052 & 4.028 & 3.911 & 3.810 & 3.666 & 3.548 \\ 
&  & TIGS (ours)      & \textbf{3.860} & \textbf{3.912} & \textbf{3.601} & \textbf{3.567} & \textbf{3.268} & \textbf{3.181} \\
\cmidrule(lr){2-9}
& \multirow{9}{*}{BLEU}
  & Template       & 0.727  & 0.815 & 0.581 & 0.687 & 0.508 & 0.559  \\
& & Seq2Seq-f      & 0.779  & 0.842 & 0.629 & 0.704 & 0.536 & 0.576  \\
& & Seq2Seq-b      & 0.774  & 0.835 & 0.623 & 0.702 & 0.534 & 0.576  \\
& & Seq2Seq-f+b    & 0.789  & 0.844 & 0.635 & 0.705 & 0.538 & 0.577  \\
& & BiRNN-BiBS     & 0.776  & 0.836 & 0.625 & 0.702 & 0.533 & 0.575  \\ 
& & BiRNN-GSN      & 0.802  & 0.848 & 0.648 & \textbf{0.707} & 0.541 & \textbf{0.579}  \\
& & Mask-Seq2Seq   & 0.785  & 0.843 & 0.635 & 0.705 & 0.537 & 0.577  \\
& & Mask-Self-attn & 0.790  & 0.845 & 0.640 & 0.706 & 0.539 & \textbf{0.579}  \\ 
& & TIGS (ours)    & \textbf{0.805} & \textbf{0.850} & \textbf{0.650} & \textbf{0.707} & \textbf{0.542} & \textbf{0.579}  \\

\midrule
\multirow{19}{*}{APRC}
& \multirow{8}{*}{NLL}
&   Seq2Seq-f      & 3.554 & 3.129 & 3.687 & 2.650 & 3.068 & 2.122  \\
& & Seq2Seq-b      & 3.694 & 3.215 & 4.039 & 2.826 & 3.494 & 2.349  \\
& & Seq2Seq-f+b    & 3.354 & 3.002 & 3.515 & 2.553 & 2.962 & 2.045  \\
& & BiRNN-BiBS     & 2.999 & 3.001 & 2.943 & 2.759 & 2.733 & 2.456  \\ 
& & BiRNN-GSN      & 2.969 & 2.967 & 2.907 & 2.515 & 2.628 & 2.012  \\
& & Mask-Seq2Seq   & 3.080 & 2.983 & 2.951 & 2.567 & 2.472 & 2.088  \\
& & Mask-Self-attn & 3.002 & 2.946 & 2.847 & 2.551 & 2.448 & 2.085 \\ 
& & TIGS (ours)      & \textbf{2.831} & \textbf{2.857} & \textbf{2.722} & \textbf{2.394} & \textbf{2.451} & \textbf{1.913} \\
\cmidrule(lr){2-9}
& \multirow{9}{*}{BLEU}
  & Template       & 0.503 & 0.692 & 0.127 & 0.432 & 0.009 & 0.182  \\
& & Seq2Seq-f      & 0.781 & 0.897 & 0.623 & 0.881 & 0.682 & 0.879 \\
& & Seq2Seq-b      & 0.779 & 0.896 & 0.616 & 0.872 & 0.683 & 0.864 \\
& & Seq2Seq-f+b    & 0.812 & 0.905 & 0.658 & 0.887 & 0.703 & \textbf{0.884} \\
& & BiRNN-BiBS     & 0.867 & 0.896 & 0.715 & 0.869 & 0.740 & 0.856 \\ 
& & BiRNN-GSN      & 0.879 & 0.904 & 0.751 & 0.884 & 0.736 & 0.882 \\ 
& & Mask-Seq2Seq   & 0.860 & 0.900 & 0.750 & 0.856 & 0.754 & 0.835 \\
& & Mask-Self-attn & 0.878 & \textbf{0.914} & \textbf{0.778} & 0.882 & \textbf{0.778} & 0.870 \\
& & TIGS (ours)      & \textbf{0.883} & 0.911 & 0.774 & \textbf{0.889} & 0.769 & 0.878 \\
\bottomrule
\end{tabular}
\vspace{5pt}
\caption{BLEU and NLL results.}
\label{tab:nll_bleu_res}
\end{center}
\vspace{-10pt}
\end{table*}

\subsection{Metrics}
Following \newcite{Sun2017}, we compare methods on standard sentence-level metric BLEU scores (4-gram) \cite{papineni2002bleu} which considers the correspondence between the ground truth and the complete sentences.
However, such a metric also has some deficiencies in text infilling tasks.
For example, given two complete sentences with only one word different, the sentence level statistics of them may be quite similar, whereas a human can clearly tell which one is most natural.
Moreover, given a template, there may be several reasonable ways to fill in the blanks. 
For example, given a template, ``i \underline{\quad} this book, highly recommend it'', it is reasonable to fill the word ``love'' or ``like'' in the blank.
However, since there is only one ground truth, the BLEU scores of these two complete sentences are quite different.
We find that this issue is more severe for the unconditional text filling task which has fewer restrictions, leading to more ways of filling in the blanks.

Therefore, for the unconditional text filling task (APRC), instead of calculating the BLEU score with only the ground truth as the reference, we also follow \newcite{Yu2016SeqGAN} and use 10,000 sentences which are randomly sampled from the test set as references to calculate BLEU scores to evaluate the fluency of the complete sentences. 

Besides BLEU scores, we conduct a model-based evaluation. We train a conditional LM for each task (unconditional LM for APRC task) and use its NLL to evaluate the quality of the complete sentence $y^*$given the input $x$. 

\subsection{Results}
The BLEU (the higher the better) and NLL (the lower the better) results are shown in Table \ref{tab:nll_bleu_res}.
Generally, we find that bidirectional methods (BiRNN-BiBS, BiRNN-GSN, and Seq2Seq-f+b) outperform unidirectional ones (Seq2Seq-f and Seq2Seq-b) in most cases.
The model-based methods (Mask-Seq2Seq and Mask-Self-attn) perform well on unconditional text infilling task (APRC), but slightly poorly on conditional text infilling tasks (Dialog and Poetry).
In line with the evaluation results in \newcite{zhu2019text}, the Mask-Self-attn performs consistently better than Mask-Seq2Seq. 
It has also achieved the highest BLEU score in some cases of unconditional text infilling tasks.
However, in most cases of conditional text infilling tasks, the proposed method performs better than Mask-Self-attn.

Since the goal of the proposed method TIGS is to find the complete sentence with minimal NLL by utilizing gradient information. 
As expected, it achieves the lowest NLL in all cases of all tasks.
Also, the BLEU scores of TIGS is highest in most cases of conditional text infilling tasks, while BiRNN-GSN and BiRNN-BiBS provide comparable performance.
Although TIGS is used in RNN-based seq2seq model, it still achieves very competitive BLEU results on unconditional text infilling task compare with Mask-Self-attn.

\begin{table}[h]
\begin{center}
\begin{tabular}{lccc}
 \hline
 Methods & Dialog & Poetry & APRC \\
 \hline 
BiRNN-BiBS & 1.524 & 1.478 & 1.558 \\ 
BiRNN-GSN & 2.979 & 2.675 & 2.261  \\  
Mask-Self-attn & 2.270 & 2.727 & 3.042  \\ 
TIGS & \textbf{3.226} & \textbf{3.120} & \textbf{3.137} \\
 \hline
 \end{tabular}
\end{center}
\vspace{-10pt}
\caption{Human evaluation resutls}
\label{tab:human}
\vspace{-10pt}
\end{table}

\subsection{Human Evaluation}
We also conduct the human evaluation to further compare TIGS, BiRNN-BiBS, BiRNN-GSN, and Mask-Self-attn. Following the setting in \newcite{zhu2019text}, we collect generations of each of the four methods on 50 randomly-selected test instances. Then we launch a crowd-sourcing online study, asking 10 evaluators to rank the generations. The method with the best generation receives a score of 4, and the other three methods receive scores of 3, 2, and 1 according to the rank, respectively. The results are shown in Table \ref{tab:human}.
We can see that TIGS consistently outperforms all baselines.

\begin{figure}[t] 
   \centering
   \includegraphics[width=3.0in]{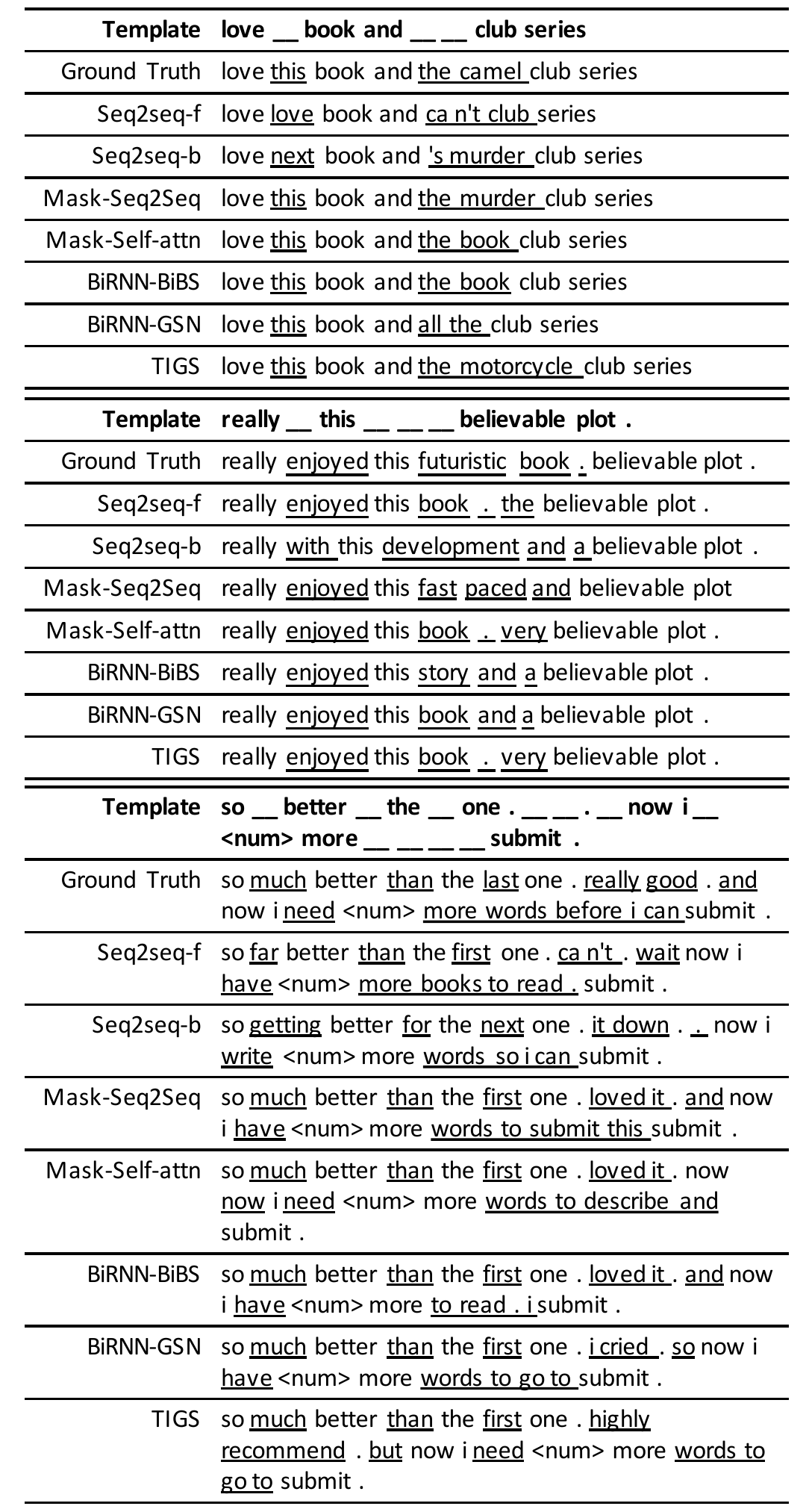}
      \vspace{-15pt}
   \caption{Example outputs of different methods on APRC task.}
      \vspace{-10pt}
   \label{fig:example_aprc}
\end{figure}

\begin{figure}[h] 
   \centering
   \includegraphics[width=3.0in]{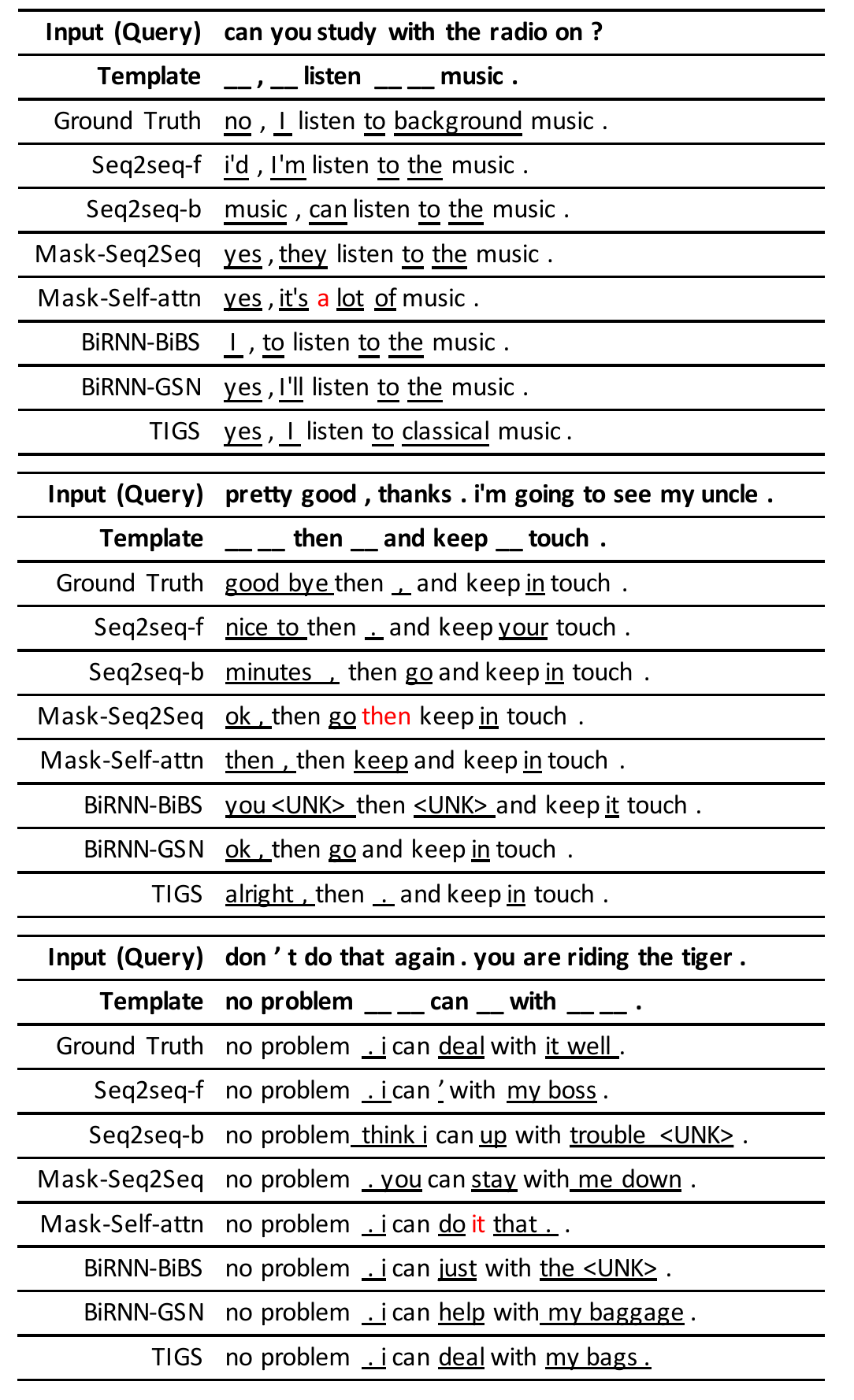}
      \vspace{-15pt}
   \caption{Example outputs of different methods on Daily task.}
      \vspace{-10pt}
   \label{fig:example_daily}
\end{figure}

\subsection{Samples and Analysis}
Figure \ref{fig:example_aprc} and \ref{fig:example_daily} show some qualitative examples of APRC and Dialog tasks.
Because the inability of Seq2Seq-f and Seq2Seq-b to reason about the past and future simultaneously. 
We can see that Seq2Seq-f and Seq2Seq-b usually generate sentences that do not satisfy grammatical rules and are not fluent.
Seq2Seq-f struggle to reason about word transitions on the forward side of the blank, so the words filled in by Seq2Seq-f usually abruptly clash with existing words behind the blank. 
Similarly, the words filled in by Seq2Seq-b usually abruptly clash with existing words before the blank.

BiRNN-BiBS makes assumption that $P(y_t|\vy_{1:t-1},\vy_{t+1:m},\vx)=P_{\overrightarrow{\mbox{\tiny URNN}}}(y_t|\vy_{1:t-1},\vx) \cdot P_{\overleftarrow{\mbox{\tiny URNN}}}(y_t|\vy_{t+1:m},\vx)$. 
This assumption may cause some sentences generated by BiRNN-BiBS are non-smooth or unreal.
For example, in the top instance, the BiRNN-BiBS generates a non-smooth sentence ``\underline{i}, \underline{to} listen \underline{to} \underline{the} music''. 
At the third time-step, because both $P_{\overrightarrow{\mbox{\tiny URNN}}}(y_3=$``to''$| y_{4:m}=$``listen \underline{to} \underline{the} music''$,\vx)$ and $P_{\overleftarrow{\mbox{\tiny URNN}}}(y_3=$``to''$| y_{1:2}=$``\underline{i},''$,\vx)$ are relatively large, resulting in this blank being filled with an inappropriate word ``to'' by BiRNN-BiBS. 
However, $P(y_3=$``to''$|y_{1:2}=$``\underline{i},''$, y_{4:m}=$``listen \underline{to} \underline{the} music''$,\vx)$ should be lower.
In addition, we find that BiRNN-BiBS tends to use the unknown token ``$<$unk$>$'' to fill the blanks compared to other methods.
The reason we analyze may be that sometimes both $P_{\overrightarrow{\mbox{\tiny URNN}}}(y_t$=``$<$unk$>$''$|\vy_{1:t-1},\vx)$ and $P_{\overleftarrow{\mbox{\tiny URNN}}}(y_t$=``$<$unk$>$''$|\vy_{t+1:m},\vx) $ would be relatively large.

As for Mask-Seq2Seq and Mask-Self-attn, although they directly take the template $\vy^\mathbb{B}$ as an additional input and are trained with data in fill-in-the-blank format. 
We experimentally found that the generalization ability of these models is still limited, especially for conditional text infilling tasks.
In the Dialog task, $21\%$ and $16\%$ of the samples generated by Mask-Self-attn and Mask-Seq2Seq with beam search could not even reconstruct the template (see Figure \ref{fig:example_daily}).\footnote{For BLEU and NLL evaluation, we force them to reconstruct the template during inference.}

Because the BiRNN-GSN fills the blank from the probability $P_{\rm \tiny BiRNN}(y_t|\vy_{1:t-1},\vy_{t+1:m},\vx)$, and the proposed method filling the blank directly with the gradient $\nabla_{\hat{\vy}_t^{emb}}\mathcal{L}(\vx,\vy^*)$.
Both of them have the ability to reason about the past and future simultaneously without any unrealistic assumptions. We can see that the complete sentences generated by them are better than all other algorithms.
However, BiRNN-GSN uses the bidirectional structure as the decoder, which makes it challenging to apply to most sequence generative models, but the proposed method is gradient-based, which can be broadly used in any sequence generative models.

\section{Conclusions}
\label{sec:conclusions}
In this paper, we propose a general inference algorithm for text infilling. To the best of our knowledge, the method is the first inference algorithm that does not require any modification or training of the model and can be broadly used in any sequence generative model to solve the fill-in-the-blank tasks.
We compare the proposed method and several strong baselines on three text infilling tasks with various mask ratios and different mask strategies. 
The results show that the proposed method is an effective and efficient approach for fill-in-the-blank tasks, consistently outperforming all baselines. 

\section*{Acknowledgment}
This work is supported by  the National Key R\&D Program of China under contract No. 2017YFB1002201, the National Natural Science Fund for Distinguished Young Scholar (Grant No. 61625204), and partially supported by the State Key Program of National Science Foundation of China (Grant Nos. 61836006 and 61432014). 

\bibliographystyle{acl_natbib}
\bibliography{filling}
\end{document}